\def\year{2020}\relax
\documentclass[letterpaper]{article} 
\usepackage{aaai20}  
\usepackage{times}  
\usepackage{helvet} 
\usepackage{courier}  
\usepackage[hyphens]{url}  
\usepackage{graphicx} 
\usepackage{graphicx}  
\usepackage{subcaption} 
\usepackage{algorithm}
\usepackage{algorithmic}
\usepackage{booktabs}
\usepackage{multirow}
\usepackage{amsmath}
\usepackage{amssymb}
\usepackage{mathrsfs}

\newcommand{\citet}[1]{\citeauthor{#1}~\shortcite{#1}}
\newcommand{\citep}{\cite}

\title{Urban2Vec: Incorporating Street View Imagery and POIs for Multi-Modal Urban Neighborhood Embedding}
\author{
Zhecheng Wang,\textsuperscript{\rm 1}\thanks{Equal contribution} 
Haoyuan Li,\textsuperscript{\rm 2}\footnotemark[1]
Ram Rajagopal\textsuperscript{\rm 1}\\
\textsuperscript{\rm 1}Stanford University,
\textsuperscript{\rm 2}
Renmin University of China\\
zhecheng@stanford.edu, 2016201695@ruc.edu.cn, ramr@stanford.edu
}

\begin{document}

\maketitle

\begin{abstract}
Understanding intrinsic patterns and predicting spatiotemporal characteristics of cities require a comprehensive representation of urban neighborhoods. Existing works relied on either inter- or intra-region connectivities to generate neighborhood representations but failed to fully utilize the informative yet heterogeneous data within neighborhoods.
In this work, we propose Urban2Vec, an unsupervised multi-modal framework which incorporates both street view imagery and point-of-interest (POI) data to learn neighborhood embeddings. 
Specifically, we use a convolutional neural network to extract visual features from street view images while preserving geospatial similarity. Furthermore, we model each POI as a bag-of-words containing its category, rating, and review information.
Analog to document embedding in natural language processing, we establish the semantic similarity between neighborhood (``document'') and the words from its surrounding POIs in the vector space. By jointly encoding visual, textual, and geospatial information into the neighborhood representation, Urban2Vec can achieve performances better than baseline models and comparable to fully-supervised methods in downstream prediction tasks. Extensive experiments on three U.S. metropolitan areas also demonstrate the model interpretability, generalization capability, and its value in neighborhood similarity analysis.
\end{abstract}

\section{Introduction}
More than 50\% of the world population live in urban areas and this share is projected to be 68\% by 2050 \cite{owidurbanization}.  
Generally, an urban area consists of a large number of spatially-distributed neighborhoods. Each neighborhood is a complex mixture of multiple components including its physical environment, local business, and the people living there. Representing and understanding the characteristics and dynamics of urban neighborhoods are essential for various downstream tasks, such as urban planning, business model development, and social welfare improvement. Survey is a common but costly method to uncover neighborhood characteristics. For example, U.S. Census Bureau spends \$250 million per year on the American Community Survey (ACS) which collects demographic statistics of different neighbourhoods \cite{gebru2017using}.
Data gathered through such time-consuming surveys can lag behind real-world urban change particularly in regions where the resources for data collection are limited. To provide decision makers with timely yet comprehensive information, we need a time-saving, cost-efficient, and general approach to characterize urban areas.

Since the intrinsic characteristics of urban neighborhoods are well embedded in their physical appearance, business activities, and even geo-tagged text information, recent data-driven methods have leveraged these widely-available data to predict or interpret the demographic and socioeconomic patterns. For example, point-of-interest (POI) categories have been used to infer region functions \cite{yuan2012discovering}; text data from Twitter has been leveraged to predict income, occupation, and unemployment rates \cite{antenucci2014using,aletras2018predicting}; imagery data, including street view images and satellite images, have been utilized to predict perceived safety, house price, and demographic makeup  \cite{naik2014streetscore,law2018take,gebru2017using}. 
However, these methods are all based on task-specific supervised learning. 
Representations learned for one task are not necessarily useful in other tasks. Furthermore, directly applying supervised learning on high-dimensional raw data, such as images, requires massive labeled data for training, which is impractical in the cases when labels are sparse.

Learning multipurpose representations of urban neighborhoods without requiring massive amounts of labeled data has led to research that represents neighborhoods as vectors while preserving the associations with urban attributes. Analog to word embeddings in NLP  \cite{mikolov2013distributed,pennington2014glove} and  graph embeddings in social network analysis \cite{perozzi2014deepwalk,grover2016node2vec}, the key idea in neighborhood embeddings is to define a ``similarity'' metrics between objects, capture it from data, and encode it into vector representations. For example, \citet{zhang2017regions} and \citet{yao2018representing} extract ``similarity'' between regions from urban mobility flow, and optimize the neighborhood embedding to represent the neighborhood function via either skip-gram or matrix factorization techniques. 
By contrast, \citet{wang2018learning} obtained the region embedding with the ``similarity'' measurement derived from the POI categories and inter-POI connectivity inside each neighborhood. 
Besides intra-region POI information, \citet{fu2019efficient} further incorporated the inter-region proximity based on both distance and POI-derived functionality to improve the embedding.
However, all these methods either relied on connectivities between neighborhoods regardless of components inside neighborhoods or included only a single type of component such as POI.
In reality, neighborhoods are complex systems involving the local natural and built environment, population, businesses and their interconnections.
Representations can only capture these relationships if different \emph{modalities} of granular data are combined. To the best of our knowledge, no existing work has fully utilized the rich and multi-modal information inside neighborhoods and incorporated them together to generate the neighborhood representations.

To bridge this gap, we propose \textbf{Urban2Vec}, an \textit{unsupervised} and \textit{multi-modal} method for learning compact yet comprehensive representations of urban neighborhoods by incorporating heterogeneous data associated with them. We model each neighborhood as a ``container'' consisting of inter-correlated components including its physical environment, business activities, and population. 
While population demographics are expensive to obtain and update, features of physical environment and business activities can be well captured in the widely-available and frequently-updated street views and POI data, respectively. Based on this idea, we develop a multi-stage approach to integrate information from street view images and POIs to generate the neighborhood representations. For street view images, visual features are extracted using convolutional neural network (CNN) while preserving geospatial correlation. For POIs, instead of using POI statistics such as number of POIs in different categories, we model each POI as a bag-of-words by collecting the textual information on its price, rating, and reviews. A major advantage of such modeling is that the semantic correlations lying behind words can be captured and projected into the vector space. 
Specifically, our contribution is three-fold:

\begin{itemize}
\item We develop a multi-modal and multi-stage framework to generate neighborhood representations integrating both image and textual data inside neighborhoods. 
Unlike previous multi-modal embedding techniques that aim at establishing inter-correlations between different modalities of objects, our major goal is to establish the correlation between the "container" (neighborhood) and its inside objects.
In this paper, the inside objects are street view images and POI data, but the framework can be extended to other data such as satellite images and geo-tagged posts.                             
\item We conduct experiments on three metropolitan areas: Bay Area, New York, and Chicago. For demographic and socioeconomic attribute prediction task, Urban2Vec achieves performances better than baseline methods and comparable to fully-supervised methods yet with better generalization capability.
\item We map the neighborhoods of New York and Chicago into the same vector space, and illustrate how the embeddings obtained through Urban2Vec can be used to draw similarities between neighborhoods across different cities.
\end{itemize}

\section{Problem Statement}
\subsubsection{Definition 1 \textit{(Urban Neighborhood)}.} \textit{A city or a metropolitan area can be represented by a set of urban neighborhoods $\mathcal{R} = \{r_1, r_2, ..., r_N\}$. Each urban neighborhood $r_i$ contains a set of street view images $\mathcal{S}_i = \{s_{i1}, s_{i2}, ..., s_{i M_i}\}$ taken inside $r_i$, and a set of POIs $\mathcal{P}_i = \{p_{i1}, p_{i2}, ..., p_{i O_i}\}$ located in $r_i$.}

\subsubsection{Remark 1.} Street view images capture the physical environment in a neighborhood, while POIs, including various types of restaurants, stores, schools, etc., represent the business activities inside the neighborhood. 
Here we assume the GPS coordinates of both street views and POIs are known.

\subsubsection{Definition 2 \textit{(POI Textualization)}.} \textit{A POI $p$ can be textualized as a bag of words $\{t_1, t_2, ..., t_q\}$, with the words extracted from $p$'s categories, rating, price and customer reviews. Then, the POI set of neighborhood $r_i$ can also be represented by a collection of POI words $\mathcal{T}_i = \{t_{i1}, t_{i2}, ..., t_{i H_i}\}$, which is obtained by merging the bags-of-words of all POIs in $r_i$.}

\subsubsection{Remark 2.} Rather than generate POI embeddings as an intermediate step, we merge bags-of-words of POIs to obtain a larger bag-of-words for each neighborhood in order to simplify the training process. 
Furthermore, such bag-of-words modeling can be easily generalized to any other geo-tagged textual data, such as Twitter and Facebook posts. 
The details of constructing $\mathcal{T}_i$ for each neighborhood will be introduced in Methodology. Here we formulate the problem to investigate as follows:

\subsubsection{Definition 3 \textit{(Urban Neighborhood Embedding)}.} \textit{Given a set of neighborhoods $\mathcal{R}$, together with street view collection $\mathcal{S}_i$ and POI word collection $\mathcal{T}_i$ inside each neighborhood $r_i$, the aim is to  learn a vector representation $z_i \in \mathbb{R}^d$ for each $r_i$, where $d$ is the uniform dimension for all $r_i$ in $\mathcal{R}$.}

\section{Methodology}
In this section, we introduce the multi-stage framework, Urban2Vec, for urban neighborhood embedding. We first present a framework overview, followed by the detailed steps for incorporating street view images and POI data.

\subsection{Framework Overview}
Each neighborhood $r_i$ is modeled as a ``container'' consisting of a street view collection $\{s_{ij}\}_{1 \leq i \leq N, 1 \leq j \leq M_i}$ and a POI bag-of-words $\{t_{ij}\}_{1 \leq i \leq N, 1 \leq j \leq H_i}$. Besides the embedding $z_i \in \mathbb{R}^d$ for each $r_i$, we also generate embedding $x_{ij} \in \mathbb{R}^d$ for each street view $s_{ij}$, and embedding $y_{ij} \in \mathbb{R}^d$ for each POI word $t_{ij}$. 
To incorporate both semantic and geospatial information extracted from street views and POIs into neighborhood representations, we seek to establish relationships between $\{z_i\}_{1 \leq i \leq N}$ and $\{x_{ij}\}_{1 \leq i \leq N, 1 \leq j \leq M_i}$ as well as $\{y_{ij}\}_{1 \leq i \leq N, 1 \leq j \leq H_i}$ in the vector space. 
Specifically, for each urban neighborhood $r_i$, we define its street view ``context'' as the collection of street views located in that neighborhood, which is $\mathcal{S}_i$, and its POI word ``context'' as the collection of words from POIs in that neighborhood, which is $\mathcal{T}_i$.
Our basic assumption is that a neighborhood should be highly correlated with the street views and POI words inside its ``context'', while less correlated with those outside its ``context''. To this end, we aim to minimize the distances of neighborhood with both its street view ``context'' and POI word ``context'' in the vector space: 

\begin{equation}\label{eq1}
L_{nb\text{-}sv}= \sum_{i=1}^N \sum_{j=1}^{M_i} \text{dist}(z_i, x_{ij}) \quad s_{ij} \in \mathcal{S}_i
\end{equation}
\begin{equation}\label{eq2}
L_{nb\text{-}poi}= \sum_{i=1}^N \sum_{j=1}^{H_i} \text{dist}(z_i, y_{ij}) \quad t_{ij} \in \mathcal{T}_i
\end{equation}
where ``dist'' is a distance function in the vector space. Here we use Euclidean distance. 
To obtain street view embedding $\{x_{ij}\}$, we apply a CNN on raw images $\{s_{ij}\}$ to extract visual features and project down the dimensionality. By contrast, the embedding of a POI word is retrieved from an embedding matrix $Y \in \mathbb{R}^{|C| \times d}$, where $|C|$ is the size of the POI word corpus. 
Minimizing $L_{nb\text{-}sv}$ involves much more parameters to train than minimizing $L_{nb\text{-}poi}$, which can make their paces of convergence difficult to be concordant. Therefore simultaneously minimizing them can yield unsatisfactory results (See performance comparison in Appendix A.1.2).
Instead, we propose an approach to optimize them separately in two stages. In the first stage, we minimize $L_{nb\text{-}sv}$ leveraging both geospatial relationship between neighborhood and street views, and that between street view and street view. 
In the second stage, we minimize $L_{nb\text{-}poi}$ to further improve the neighborhood embedding by incorporating textual information of POIs.

\subsection{Incorporating Street View Imagery}

\begin{figure}[t]
\begin{center}
\includegraphics[width=0.95\linewidth]{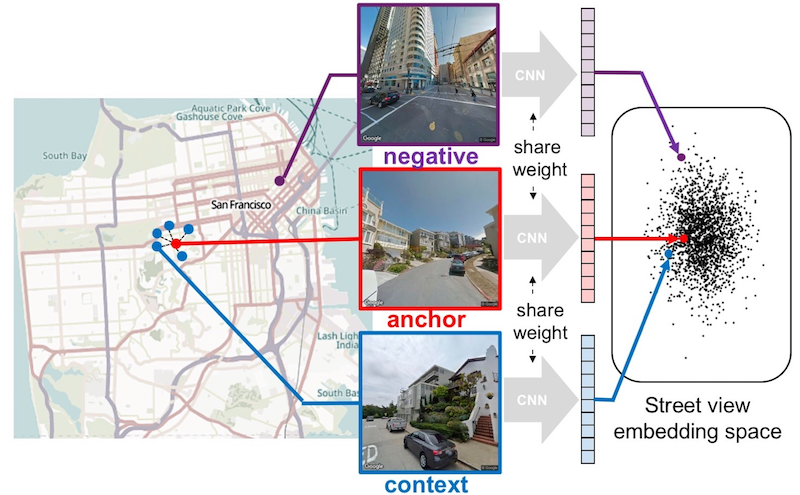}
\end{center}
\caption{Street view embedding with negative sampling. The anchor image is mapped into closer position to context image than negative image in the vector space.}
\label{fig:sv}
\end{figure}

Each street view image can appear in the context of only one neighborhood, thus directly minimizing 
$L_{nb\text{-}sv}$ will generate multiple \textit{isolated} agglomerations in the vector space, where each agglomeration is formed with a neighborhood embedding surrounded by the embeddings of its contextual street views. This is contrary to the fact that neighborhoods are spatially correlated with each other rather than isolated. To tackle this issue, we further decompose the process by firstly minimizing the distance between each street view and its contextual street views: 

\begin{equation}\label{eq3}
L_{sv\text{-}sv}= \sum_{k} \sum_{\quad x_{c} \in \mathcal{N}_{k}} \text{dist}(x_{k}, x_{c}) 
\end{equation}
and then minimizing $L_{nb\text{-}sv}$ while keeping all $x_{k}$ fixed. Here $\mathcal{N}_{k}$ denotes the ``context'' of street view $s_{k}$.

\subsubsection{Triplet loss.} To optimize $L_{sv\text{-}sv}$, we need to firstly answer the question: What can be the ``context'' of a street view image? 
In satellite image tile embedding \cite{jean2019tile2vec}, the ``context'' is defined as the geographic neighbors of a tile, based on the assumption that tiles that are spatially close should have higher semantic similarity, while tiles far apart tend to have less. 
In our work, we extend this assumption to street view images: Street views with small geographic distance are more likely to share common semantics than those with large distance. Therefore, we also define the ``context'' of a street view image as its geospatial neighbors. 
Specifically, to optimize $L_{sv\text{-}sv}$, we train the CNN on a collection of triplets $(s_a, s_c, s_n)$, where $s_a$ is the anchor image, $s_c$ is an image inside $s_a$'s context, and $s_n$ is a negative sample outside $s_a$'s context (See Figure \ref{fig:sv}). To enforce such relative similarity in the vector space, we minimize the triplet loss to enlarge the margin between anchor-negative distance and anchor-context distance:

\begin{equation}\label{eq4}
L_{sv\text{-}sv}^{tri}(x_a, x_c, x_n) = [m+||x_a - x_c||_2 - ||x_a - x_n||_2]_{+}
\end{equation}
where $[\cdot]_{+}$ is a rectifier and $||\cdot||_2$ is the Euclidean distance. 
Margin $m$ is used to prevent infinitely large difference between these two distances.
$x_a$, $x_c$, and $x_n$ are the embeddings of $s_a$, $s_c$, and $s_n$, respectively, which are outputted by a shared CNN, i.e., $x\cdot = f_\theta (s\cdot)$ where $f_\theta$ denotes a CNN parameterized by $\theta$. Note that $L_{sv\text{-}sv}^{tri}$ can be recognized as an implementation of $L_{sv\text{-}sv}$ with negative sampling. 

\subsubsection{Triplet sampling. } Rather than use absolute geographic distance to determine the ``context'', we define the ``context'' of an anchor image as its $K$ nearest images. This is because street view images can be densely distributed in some regions while sparse in others. Using absolute geographic distance will result in highly variable context size for different anchor images. 

\subsubsection{Minimizing $L_{nb\text{-}sv}$. }
After obtaining the street view embedding $\{x_{ij}\}$ by minimizing $L_{sv\text{-}sv}^{tri}$, we then minimize $L_{nb\text{-}sv}$ while keeping all $x_{ij}$ fixed. We do not apply negative sampling in this step, since the relative similarity has already been captured through the geospatial context in street view embedding. Instead, we simply take the average of all $x_{ij}$ located in neighborhood $i$ as the neighborhood embedding $z_{i}$, which is exactly the analytical solution of minimizing $L_{nb\text{-}sv}$:

\begin{equation}\label{eq5}
z_i = \frac{1}{M_i} \sum_{j=1}^{M_i} x_{ij} \quad s_{ij} \in \mathcal{S}_i
\end{equation}
This process is illustrated in Figure \ref{fig:nb}.

\subsection{Incorporating POI Textual Data}
\begin{figure}[t]
\begin{center}
\includegraphics[width=0.92\linewidth]{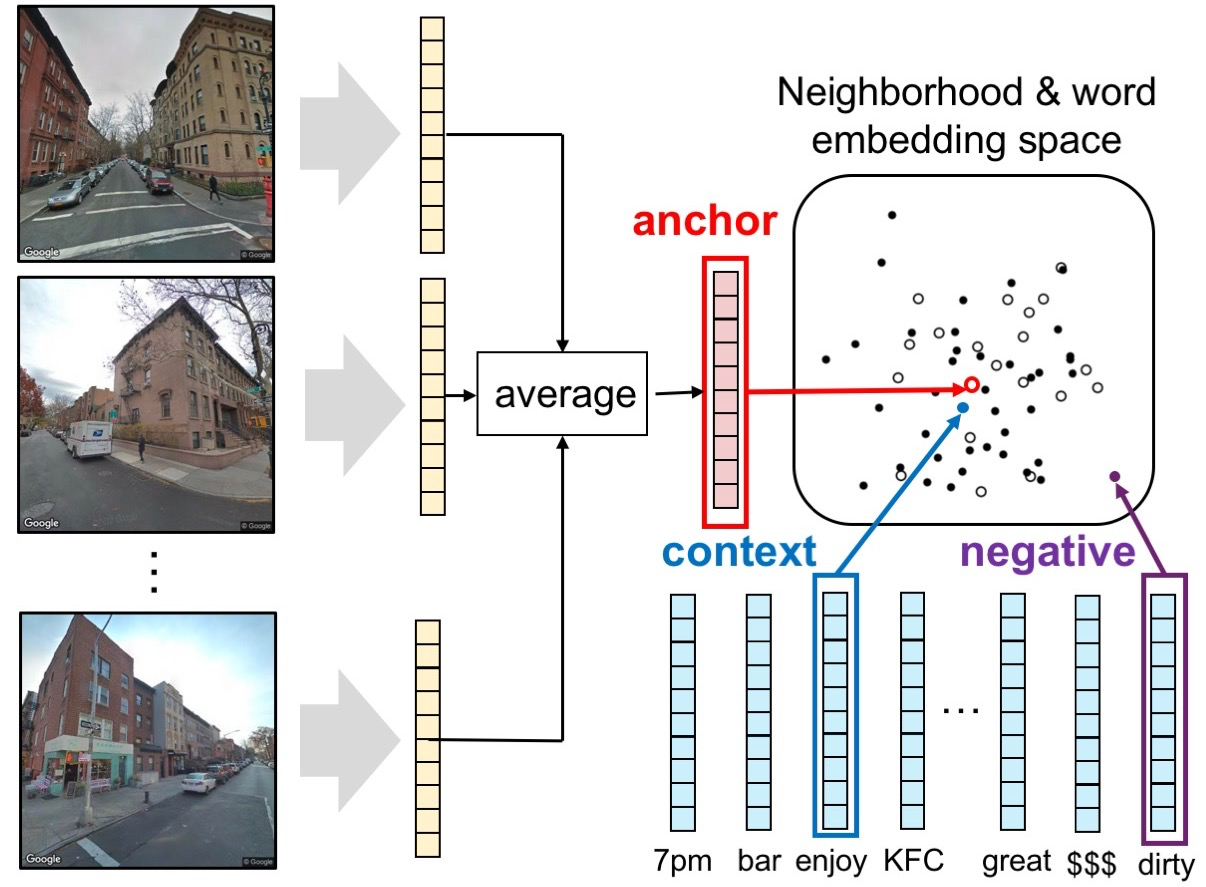}
\end{center}
\caption{Generating neighborhood embedding by aggregating street view embedding and interacting with POI word embedding.}
\label{fig:nb}
\end{figure}

In contrast to street view embedding which leverages CNN to project down the dimensionality, the vector representation of each POI word is directly retrieved from the word embedding matrix $Y$.
This is analog to document embedding \cite{le2014distributed}, where a neighborhood can be regarded as a document containing words.
To encode textual semantics regarding local business into neighborhood representations, we define another triplet loss to train both POI word embedding $Y$ and neighborhood embedding $Z = \{z_i\}_{1 \leq i \leq N} \in \mathbb{R}^{N \times d}$. 
In this stage, each triplet $(r_a, t_c, t_n)$ consists of an anchor neighborhood $r_a$, a POI word $t_c$ from $r_a$'s context, and a negative sample $t_n$ outside $r_a$'s context. Their embeddings are $z_a$, $y_c$, and $y_n$, respectively.
Here we define the POI word ``context'' of $r_a$ as the collection of words obtained from POIs inside $r_a$, which is $\mathcal{T}_a$. By minimizing the triplet loss:

\begin{equation}\label{eq4}
L_{nb\text{-}poi}^{tri}(z_a, y_c, y_n) = [m'+||z_a - y_c||_2 - ||z_a - y_n||_2]_{+}
\end{equation}
we make the neighborhood $r_a$ closer to its contextual POI words than to the non-contextual POI words in the vector space. Here $m'$ is another margin value serving the same purpose as in street view embedding. Note that $L_{nb\text{-}poi}^{tri}$ can also be regarded as an implementation of $L_{nb\text{-}poi}$ with negative sampling and Euclidean distance metrics. After training $Y$ and $Z$ on the POI textual dataset, the final neighborhood embedding $Z$ incorporates the semantics from both street views and POIs inside each neighborhood. The whole process is illustrated in Figure \ref{fig:nb}.

\subsubsection{Bag-of-words construction.}
To construct the POI word ``context'' $\mathcal{T}_i$ for each neighborhood, we merge the bags-of-words of all POIs located in that neighborhood. 
For each POI, its bag-of-words includes its categories, rating, price, and words from its customer reviews. 
In our setting, a category phrase (e.g. ``shopping center'') is regarded as a single word. Each category word is prefixed with ``cat$\_$'' to make it distinguishable from the same word appearing in reviews. Duplicate words are removed from reviews.
However, we do not eliminate duplication when merging words from different POIs, as the word frequency is highly indicative of the neighborhood attributes. For example, the neighbourhood with 100 ``restaurant'' occurrences tends to be more prosperous than that with only 5 ``restaurant'' occurrences.

\subsubsection{Negative sampling.}
We apply the same negative sampling scheme as in \cite{mikolov2013distributed} based on word frequencies in the corpus. Specifically, for word $t_k$ which is outside $r_i$'s context $\mathcal{T}_i$, its possibility of being selected as negative sample is $\frac{F_k^{0.5}}{\sum_{j\in\neg \mathcal{T}_i}F_j^{0.5}}$, where $F_k$ denotes the number of times that word $t_k$ appears in the corpus.

\section{Experiments}

\begin{figure*}[t]
  \centering
  \begin{subfigure}{0.5\textwidth}
    \centering
 .  \includegraphics[width=\textwidth]{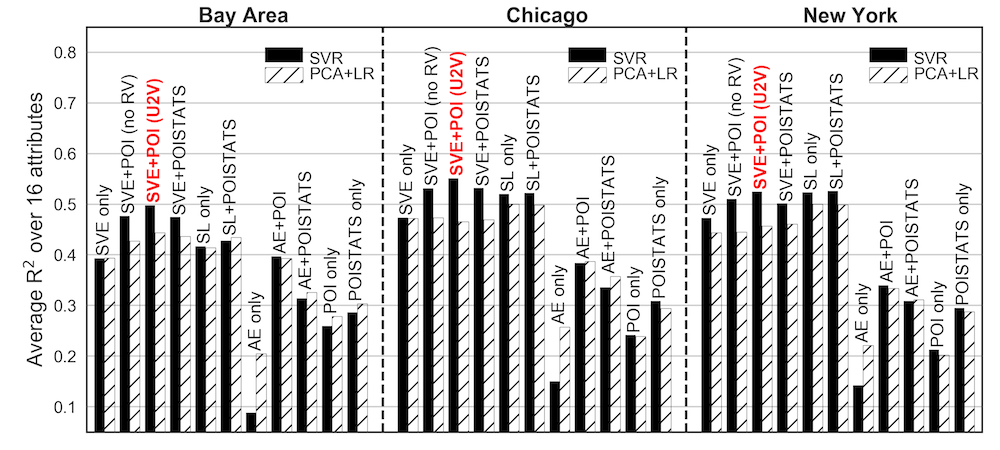}
    \caption{Average R$^2$}
    \label{fig:reg_average}
  \end{subfigure}%
  \hfill
  \begin{subfigure}{0.5\textwidth}
    \centering
 .  \includegraphics[width=\textwidth]{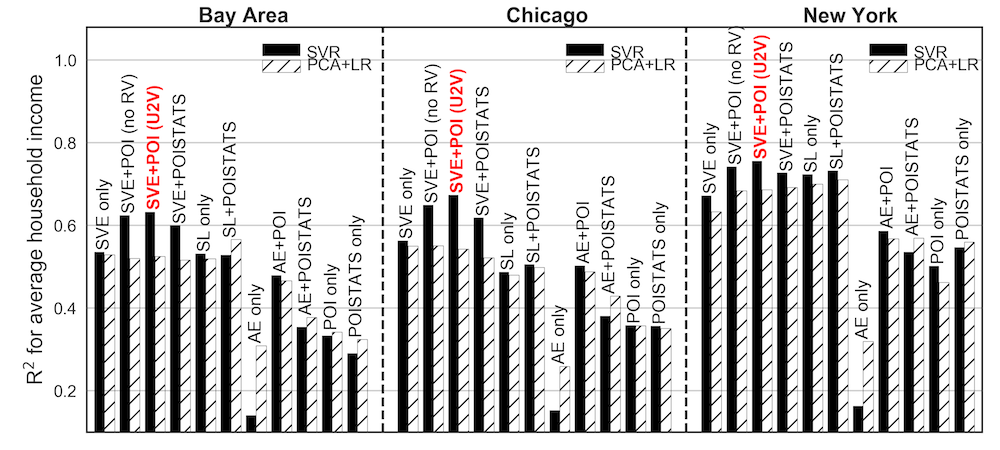}
    \caption{Average household income (\textbf{AHI})}
    \label{fig:reg_income}
  \end{subfigure}
  \hfill
  %
  \begin{subfigure}{0.5\textwidth}
    \centering
 .  \includegraphics[width=\textwidth]{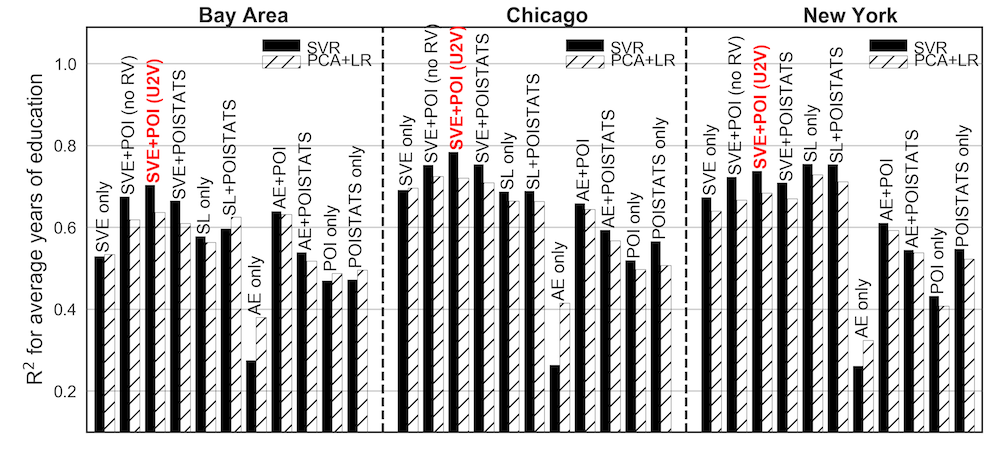}
    \caption{Average years of education (\textbf{AYE})}
    \label{fig:reg_education}
  \end{subfigure}%
  \hfill
  \begin{subfigure}{0.5\textwidth}
    \centering
 .  \includegraphics[width=\textwidth]{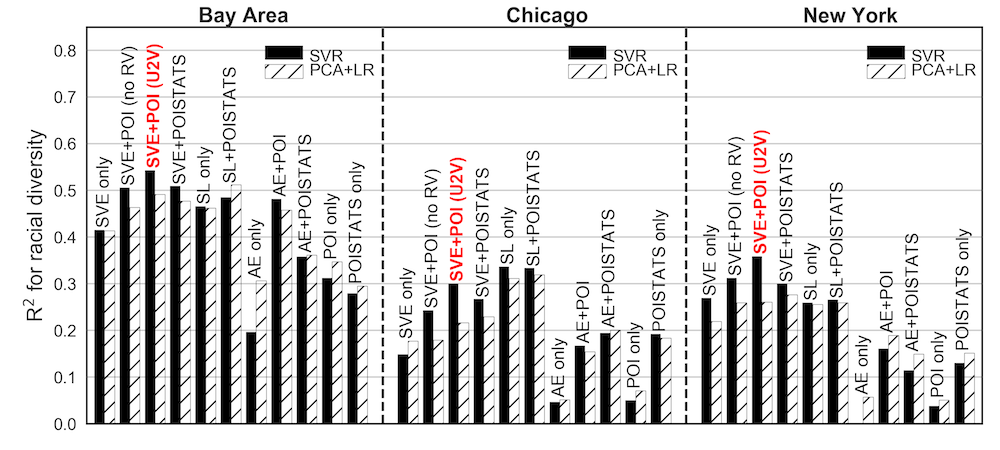}
    \caption{Racial diversity (\textbf{RD})}
    \label{fig:reg_diversity}
  \end{subfigure}%
  
  \caption{Performances in demographic attribute prediction based on both SVR (black bar) and PCA+LR (hatching bar). See abbreviation description in \textbf{Experiments-Baseline Models} section.}
  \label{fig:reg}
\end{figure*}

By conducting experiments on real-world dataset, we demonstrate: (1) the performances of using the neighborhood embedding derived by Urban2Vec for downstream prediction tasks, (2) the interpretability underlying the vector representation, and (3) the usage of the embedding for neighborhood similarity analysis. 

In this work, we use census tracts to define ``neighborhoods''. Census tracts are small and relatively homogeneous units regarding their demographics, and the census-tract-level data of these attributes is readily available, which is favorable for our model evaluation.  
Our framework is easy to be extended to other geographic units such as block groups or counties, which is a part of our future work.

\subsection{Experimental Setups}
We collect street view images and POI information in three major metropolitan areas in the US: San Francisco Bay Area, Chicago, and New York. Street view images are retrieved using Google Street View Static API. 
POI information is obtained with Yelp Fusion API.
The statistics of our datasets are shown in Table \ref{table1}.

\begin{table}[h]
\centering
\resizebox{.95\columnwidth}{!}{
\begin{tabular}{lcccc}
\hline
         & \# street views       & \# POIs   & \# unique words  & \# census tracts \\ \hline
Bay Area & 44174             & 69765 & 14279 & 1198         \\
Chicago  & 64739             & 38445 & 10013 & 1317         \\
New York & 67271             & 50697 & 11386 & 1371         \\ \hline
\end{tabular}}
\caption{Dataset statistics.}
\label{table1}
\end{table}

To obtain the street view embedding, we use context size $K=5$. The CNN used in our work is an Inception-v3 architecture \cite{szegedy2016rethinking} with a final linear layer projecting 2048-dimensional feature into the $d$-dimensional embedding space. We use $d=200$ for result analysis, and the dimension sensitivity analysis is shown in Appendix A.2. We use parameters pre-trained on ImageNet to intialize the CNN model.
In POI word embedding, we use the pretrained GloVe \cite{pennington2014glove} to initialize the embeddings of words in reviews, while embeddings of other words like price and rating are randomly initialized. 

\subsection{Baseline Models}
We compare Urban2Vec with other unimodal and multi-modal representations of urban neighborhoods. Specifically, we investigate four methods of incorporating street view images: (1) street view embedding (\textbf{SVE}) proposed in this paper, (2) convolutional autoencoder (\textbf{AE}) proposed in \cite{law2019learning}, which uses the encoder to embed street views into vector space, (3) supervised learning (\textbf{SL}), and (4) none. Note that for supervised learning, we use the same Inception-v3 architecture with demographic attributes as supervisory signals, which serves as a benchmark for prediction tasks. Moreover, we compare three methods of incorporating POI information: (1) POI word embedding (\textbf{POI}) proposed in this paper, (2) statistics of the POI categories using tf.idf scheme (\textbf{POISTATS}), and (3) none. 

Finally, we obtain 10 different neighborhood representations using their combinations: \textbf{SVE only}, \textbf{SVE+POI} (Urban2Vec, abbreviated as \textbf{U2V}), \textbf{SVE+POISTATS}, \textbf{SL only}, \textbf{SL+POISTATS}, \textbf{AE only}, \textbf{AE+POI}, \textbf{AE+POISTATS}, \textbf{POI only}, \textbf{POISTATS only}.

We do not compare \textbf{SL+POI} since the POI word corpus cannot be split into training and test sets for POI word embedding, while evaluating supervised learning methods requires strict training/test set partition to prevent overfitting. 
Besides, we also compare the Urban2Vec model excluding words from reviews, abbreviated as \textbf{SVE+POI (no RV)}.

\subsection{Experimental Results}

\begin{figure*}[t]
  \centering
  \begin{subfigure}{0.4\textwidth}
    \centering
 .  \includegraphics[width=\textwidth]{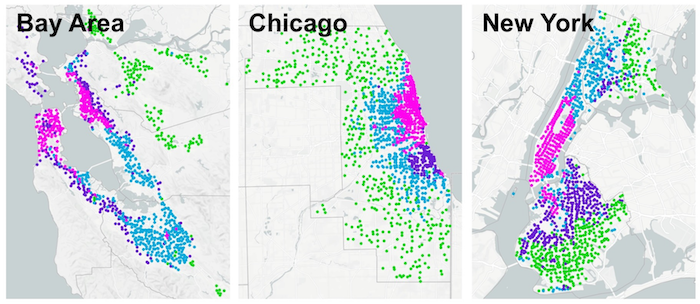}
    \caption{}
    \label{fig:clustering}
  \end{subfigure}%
  \hfill
  \begin{subfigure}{0.295\textwidth}
    \centering
 .  \includegraphics[width=\textwidth]{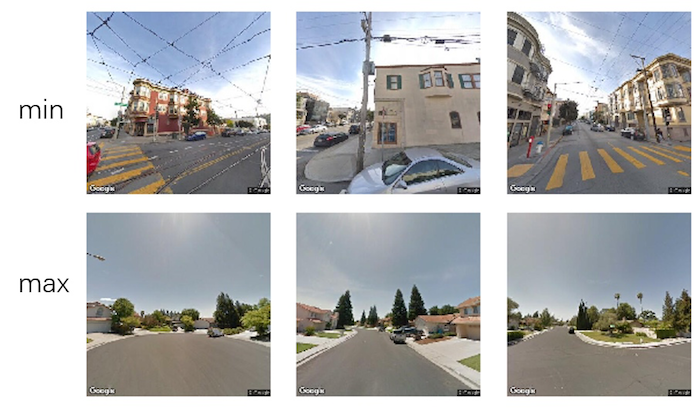}
    \caption{}
    \label{fig:pca0}
  \end{subfigure}
  \hfill
  %
  \begin{subfigure}{0.295\textwidth}
    \centering
 .  \includegraphics[width=\textwidth]{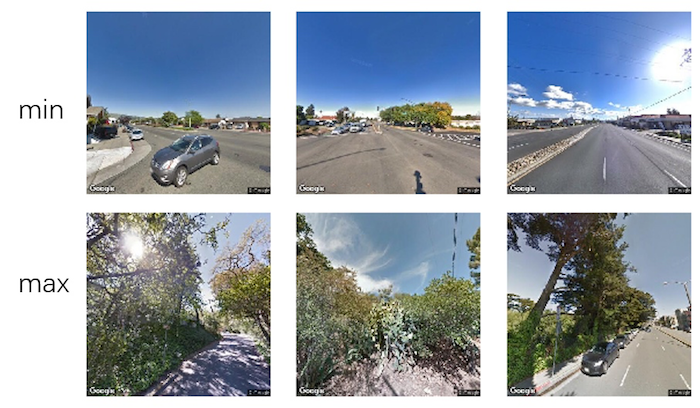}
    \caption{}
    \label{fig:pca1}
  \end{subfigure}%
  \hfill
  \caption{\textbf{(a)} $k$-means clustering on the embeddings in three areas ($k$=4). \textbf{(b)} Street views with extreme values in 1st PCA dimension. \textbf{(c)} Street views with extreme values in 2nd PCA dimension.}
  \label{fig:clustering_pca}
\end{figure*}

\subsubsection{Predicting demographic and socioeconomic attributes.}
We use neighborhood embeddings as inputs to predict their demographic attributes at census tract level. The demographic attributes are obtained from ACS and include 16 variables. 
Both Support Vector Regression (SVR) and linear regression with PCA preprocessing (PCA+LR) are used as regressors, since they are less subjective to model hyperparameters than more complex models such as Neural Network regressor.
Census tracts in each dataset are split into training (70\%), validation (15\%), and test sets (15\%). For each variable we calculate the average $R^2$ on test set over 20 random splits.
For supervised learning (\textbf{SL}), we directly use these 16 attributes for supervision during training.

Figure \ref{fig:reg} shows the prediction results of three major variables: average household income (\textbf{AHI}), average number of years of education (\textbf{AYE}), racial diversity (\textbf{RD}), together with the overall $R^{2}$ of all these 16 variables on three datasets (See the results for the rest of variables in Appendix A.3).
As is shown, when using SVR as regressor, Urban2Vec outperforms all other models in all three areas, except for predicting \textbf{AYE} in New York and predicting \textbf{RD} in Chicago on which \textbf{SL+POISTATS} performs slightly better by 0.02-0.03.
The overall $R^2$ of \textbf{SVE+POISTATS} is higher than that of \textbf{SL+POISTATS} in Bay Area but lower in New York for both regressors, while in Chicago, SVR favors \textbf{SVE+POISTATS} but PCA+LR favors \textbf{SL+POISTATS}. Such results indicate that the representation obtained with our unsupervised method can achieve comparable performances with fully-supervised method in downstream prediction tasks. For regions where granular demographic features are not readily available or out-of-date, the unsupervised learning proposed in this paper will be a good approach to obtain the neighborhood representations.
Moreover, all \textbf{SVE}-based methods have significantly higher $R^2$ than their \textbf{AE}-based counterparts. This may be because \textbf{SVE} leverage the geospatial correlation to derive representation while autoencoder only learns visual features from images themselves regardless of the relationships between images.

Regarding the POI representations, although using \textbf{POISTATS} only can yield generally better prediction performances than using \textbf{POI} only, \textbf{SVE+POI} has a $\sim$0.02 gain in average $R^2$ compared with \textbf{SVE+POISTATS} when using SVR, suggesting that in the multi-modal setting, POI word embedding is a better way to learn representations from unstructured POI data than using POI statistics as it captures the semantic nuances from the textual data.
Comparing \textbf{SVE+POI (no RV)} with \textbf{SVE+POI} shows that including customer reviews leads to a $\sim$0.02 increase in average $R^2$.
Except for \textbf{SL}-based methods, incorporating both street views and POIs  yields a remarkably better  $R^2$ than incorporating only a single mode, indicating the strength of multi-modal learning for urban neighborhood representations.

\begin{figure*}[t]
\begin{center}
\includegraphics[width=0.8\textwidth]{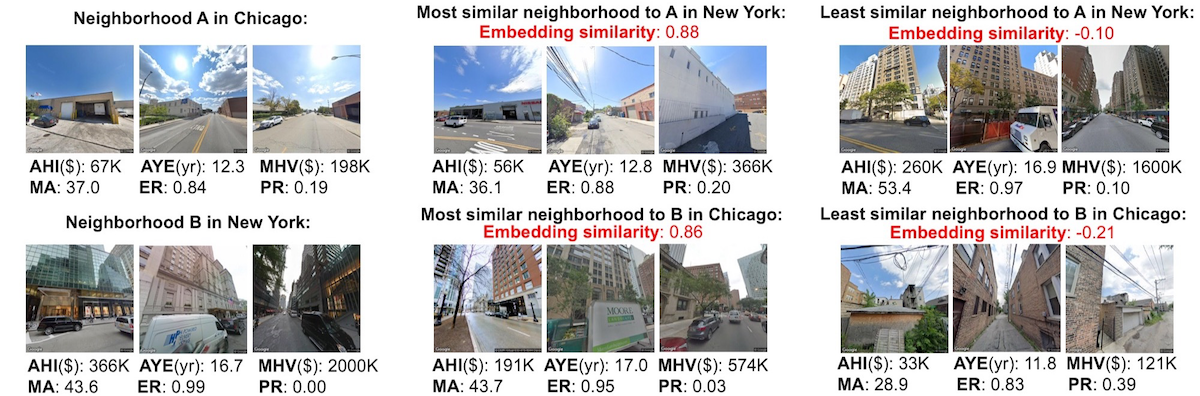}
\end{center}
\caption{Searching for the most/least similar neighborhoods in another city with Urban2Vec embedding. Abbreviation: \textbf{AHI}: average household income. \textbf{AYE}: average years of education. \textbf{MHV}: median housing unit value. \textbf{MA}: median age. \textbf{ER}: employment rate. \textbf{PR}: ratio of families below poverty level.}
\label{fig:similarity}
\end{figure*}

\begin{table}[h]
\centering
\resizebox{\columnwidth}{!}{
\begin{tabular}{lcccc}
\hline
\multicolumn{1}{l}{} & Urban2Vec & AE+POI & SL+POISTATS & demographic \\ \hline
Avg apartment sale price          & \textbf{0.487}  & 0.391           & 0.417          & 0.445     \\
Avg office sale price           & \textbf{0.443}  & 0.308           & 0.328          & 0.337   \\ \hline
\end{tabular}}
\caption{Real estate sale price prediction $R^2$.}
\label{table2}
\end{table}

We also evaluate embedding methods on predicting real estate sale price in Bay Area (Table \ref{table2}). The data is obtained from Cushman \& Wakefield and aggregated at census tract level. For \textbf{SL+POISTATS}, in order to compare its generalization capability with Urban2Vec, we still use the vector representation obtained from the demographic prediction as inputs, instead of re-training it with the real estate sale price as supervision.
Result shows that for both apartment and office sale price predictions, Urban2Vec outperforms other methods including directly using demographic attributes as inputs.
Though the performances of supervised method and Urban2Vec are comparable in demographic prediction, when transferring the learned representations to another task, Urban2Vec performs significantly better, indicating its good generalization capability.

\subsubsection{Interpreting the learned representations.}
We apply $k$-means clustering on the embeddings obtained with Urban2Vec in three areas. 
Figure \ref{fig:clustering} shows the results with $k$=4. In Bay Area, regions with higher street enclosure and denser population, such as San Francisco and Oakland, are clustered into the same group. Similar cluster also occurs in Chicago and New York (i.e. Central Chicago, most part of Manhattan). 
In each area, regions with less dense population and buildings are clustered into another group (green). 
Such clustering results align well with human intuition on the city structures of these three areas.

To uncover the semantic meaning learned in neighborhood representations, we apply PCA on the embedding matrix obtained in Bay Area, and visualize the street views and POI category words with extreme values along the directions of first and second principle components.
They are shown in Figure \ref{fig:clustering_pca} and Table \ref{table3}, respectively. 
We find that the 1st component is associated with street enclosure (average building heights divided by average width between buildings of the same street), while the 2nd component is correlated with vegetation. Interestingly, we also observe that the 2nd component has strong positive correlation with average household income and education level when we use PCA+LR to predict demographic attributes in Bay Area, suggesting that denser greenery is indicative of high income/education level in Bay Area. Moreover, POI categories such as ``gallery'' and ``winery'' have high values in the 2nd dimension, while ''discount store'' has low value (Table \ref{table3}), indicating that they are potentially associated with high income/education level and low income/education level, respectively.

\begin{table}[h]
\centering
\resizebox{\columnwidth}{!}{
\begin{tabular}{lllll}
\hline
dim1 & min & dance club  & comedy club   & popup restaurant  \\ \cline{2-5} 
            & max  & horse back riding & winery     & pool cleaning    \\ \hline
dim2 & min & water store     & discount store   &      \\ \cline{2-5} 
            & max & gallery     & bed breakfast & winery           \\ \hline
\end{tabular}}
\caption{POI category words with extreme values in 1st and 2nd PCA dimensions.}
\label{table3}
\end{table}

\subsubsection{Neighborhood similarity analysis.}

Vector representation of neighborhoods can be used to draw similarity efficiently between neighborhoods subject to geographic constraints.
Specifically, we seek to answer questions like this: Which neighborhood in Chicago is most similar to neighborhood $X$ in New York? 
To illustrate the value of Urban2Vec in addressing such problems, we train a joint embedding using Urban2Vec for all neighborhoods in New York and Chicago.
Given a source neighborhood in one city, we can compute the cosine similarity between its embedding vector and all embedding vectors in another city, and then rank the neighborhoods according to the similarity metrics. Figure \ref{fig:similarity} shows two instances of similar neighborhood search (See more instances in Appendix A.4). Neighborhood $A$ in Chicago and its most similar neighborhood in New York obtained through Urban2Vec share the similar physical appearance and demographic attributes: warehouse-like buildings, sparse vegetation, as well as medium-to-low income and employment rate. By contrast, the least similar neighborhood to $A$ in New York features tall buildings, high income/housing unit value/employment rate, and a more aging community.
Also, neighborhood $B$ in New York and its ``nearest neighbor'' in Chicago both feature tall apartments, high income/education level/housing unit values, and near-zero poverty rate, while the least similar neighborhood to $B$ in Chicago is characterized by its dilapidated buildings and opposite demographic features.
To summarize, Urban2Vec transforms the complicated urban system comparison into a simple vector comparison problem, which can facilitate commercial site selection, real estate recommendation, and urban planning.

\section{Conclusion}
In this work, we propose Urban2Vec, an unsupervised multi-modal learning framework to learn neighborhood representations by incorporating both street view and POI data. We demonstrate its high accuracy in downstream prediction tasks, interpretability underlying the embedding space, and its potential usage in neighborhood similarity analysis. Future work includes (1) incorporating other imagery, textual, and spatiotemporal data, such as mobility and social media data, to increase the information richness of neighborhood representations,  (2) using user study to evaluate neighborhood similarity ranking performance, and (3) exploring the potential usage of Urban2Vec in analyzing urban change.

\section{Related Work}
\subsubsection{Image embedding.} Unsupervised representation of images, or image embedding, aims to project high-dimensional image data into compact but informative representations. 
In terms of specific applications in urban study, \citet{law2019learning} used convolutional autoencoder followed by PCA to generate representations of both street views and street network images. However, directly applying autoencoder on images does not establish any geospatial correlation between street views at different locations. 
\citet{jean2019tile2vec} developed an algorithm to learn vector representations of satellite images by using geospatial information as a weak supervision, which provided a paradigm for learning the representation of spatially distributed image data. 

\subsubsection{Text and graph embedding.} 
In text embedding, the target is to enforce single words \cite{mikolov2013distributed} or word sequences \cite{le2014distributed} in the same context to be close in the vector space,
In graph embedding, each node is treated as a word with its ``context'' defined as a set of other nodes along a random walk \cite{perozzi2014deepwalk,grover2016node2vec}.
Inspired by them, existing neighborhood embedding methods modeled mobility flow as graphs or origin-destination pairs either within a region \cite{wang2018learning,fu2019efficient} or among regions \cite{zhang2017regions,yao2018representing}.
Compared to these works, we fully utilize the heterogeneous data in each neighborhood and incorporate multiple modalities of components to obtain the comprehensive neighborhood representations.

\subsubsection{Multi-modal embedding.} Mapping different modalities of data into same latent space has been studied before. It has been shown that images and text can be projected into a single space using neural networks \cite{wang2016learning,zhang2017regions},
so did the combination of point-cloud, text, and robot manipulation trajectories \cite{sung2017deep}, and the combination of text, location, and time \cite{zhang2017regions}.
However, instead of aiming at embedding data of different modalities into one space, the question we are interested in is how to obtain the embedding of a ``container'' (neighborhood) by integrating multiple modalities of data inside the ``container''.

\section{Acknowledgements}
We thank Amazon Web Service for offering cloud computing credits and Cushman \& Wakefield for offering the real estate data. Z. Wang was supported by the Stanford Interdisciplinary Graduate Fellowship as Satre Family Fellow.

\bibliography{6242papers}

\begin{thebibliography}{}

\bibitem[\protect\citeauthoryear{Aletras and
  Chamberlain}{2018}]{aletras2018predicting}
Aletras, N., and Chamberlain, B.~P.
\newblock 2018.
\newblock Predicting twitter user socioeconomic attributes with network and
  language information.
\newblock In {\em Proceedings of the 29th on Hypertext and Social Media},
  20--24.
\newblock ACM.

\bibitem[\protect\citeauthoryear{Antenucci \bgroup et al\mbox.\egroup
  }{2014}]{antenucci2014using}
Antenucci, D.; Cafarella, M.; Levenstein, M.; R{\'e}, C.; and Shapiro, M.~D.
\newblock 2014.
\newblock Using social media to measure labor market flows.
\newblock Technical report, National Bureau of Economic Research.

\bibitem[\protect\citeauthoryear{Fu \bgroup et al\mbox.\egroup
  }{2019}]{fu2019efficient}
Fu, Y.; Wang, P.; Du, J.; Wu, L.; and Li, X.
\newblock 2019.
\newblock Efficient region embedding with multi-view spatial networks: A
  perspective of locality-constrained spatial autocorrelations.

\bibitem[\protect\citeauthoryear{Gebru \bgroup et al\mbox.\egroup
  }{2017}]{gebru2017using}
Gebru, T.; Krause, J.; Wang, Y.; Chen, D.; Deng, J.; Aiden, E.~L.; and Fei-Fei,
  L.
\newblock 2017.
\newblock Using deep learning and google street view to estimate the
  demographic makeup of neighborhoods across the united states.
\newblock {\em Proceedings of the National Academy of Sciences}
  114(50):13108--13113.

\bibitem[\protect\citeauthoryear{Grover and
  Leskovec}{2016}]{grover2016node2vec}
Grover, A., and Leskovec, J.
\newblock 2016.
\newblock node2vec: Scalable feature learning for networks.
\newblock In {\em Proceedings of the 22nd ACM SIGKDD international conference
  on Knowledge discovery and data mining},  855--864.
\newblock ACM.

\bibitem[\protect\citeauthoryear{Jean \bgroup et al\mbox.\egroup
  }{2019}]{jean2019tile2vec}
Jean, N.; Wang, S.; Samar, A.; Azzari, G.; Lobell, D.; and Ermon, S.
\newblock 2019.
\newblock Tile2vec: Unsupervised representation learning for spatially
  distributed data.
\newblock In {\em Proceedings of the AAAI Conference on Artificial
  Intelligence}, volume~33,  3967--3974.

\bibitem[\protect\citeauthoryear{Law and Neira}{2019}]{law2019learning}
Law, S., and Neira, M.
\newblock 2019.
\newblock Learning from discovering: An unsupervised approach to geographical
  knowledge discovery using street level and street network images.
\newblock {\em arXiv preprint arXiv:1906.11907}.

\bibitem[\protect\citeauthoryear{Law, Paige, and Russell}{2018}]{law2018take}
Law, S.; Paige, B.; and Russell, C.
\newblock 2018.
\newblock Take a look around: using street view and satellite images to
  estimate house prices.
\newblock {\em arXiv preprint arXiv:1807.07155}.

\bibitem[\protect\citeauthoryear{Le and Mikolov}{2014}]{le2014distributed}
Le, Q., and Mikolov, T.
\newblock 2014.
\newblock Distributed representations of sentences and documents.
\newblock In {\em International conference on machine learning},  1188--1196.

\bibitem[\protect\citeauthoryear{Mikolov \bgroup et al\mbox.\egroup
  }{2013}]{mikolov2013distributed}
Mikolov, T.; Sutskever, I.; Chen, K.; Corrado, G.~S.; and Dean, J.
\newblock 2013.
\newblock Distributed representations of words and phrases and their
  compositionality.
\newblock In {\em Advances in neural information processing systems},
  3111--3119.

\bibitem[\protect\citeauthoryear{Naik \bgroup et al\mbox.\egroup
  }{2014}]{naik2014streetscore}
Naik, N.; Philipoom, J.; Raskar, R.; and Hidalgo, C.
\newblock 2014.
\newblock Streetscore-predicting the perceived safety of one million
  streetscapes.
\newblock In {\em Proceedings of the IEEE Conference on Computer Vision and
  Pattern Recognition Workshops},  779--785.

\bibitem[\protect\citeauthoryear{Pennington, Socher, and
  Manning}{2014}]{pennington2014glove}
Pennington, J.; Socher, R.; and Manning, C.
\newblock 2014.
\newblock Glove: Global vectors for word representation.
\newblock In {\em Proceedings of the 2014 conference on empirical methods in
  natural language processing (EMNLP)},  1532--1543.

\bibitem[\protect\citeauthoryear{Perozzi, Al-Rfou, and
  Skiena}{2014}]{perozzi2014deepwalk}
Perozzi, B.; Al-Rfou, R.; and Skiena, S.
\newblock 2014.
\newblock Deepwalk: Online learning of social representations.
\newblock In {\em Proceedings of the 20th ACM SIGKDD international conference
  on Knowledge discovery and data mining},  701--710.
\newblock ACM.

\bibitem[\protect\citeauthoryear{Ritchie and Roser}{2019}]{owidurbanization}
Ritchie, H., and Roser, M.
\newblock 2019.
\newblock Urbanization.
\newblock {\em Our World in Data}.
\newblock https://ourworldindata.org/urbanization.

\bibitem[\protect\citeauthoryear{Sung, Lenz, and Saxena}{2017}]{sung2017deep}
Sung, J.; Lenz, I.; and Saxena, A.
\newblock 2017.
\newblock Deep multimodal embedding: Manipulating novel objects with
  point-clouds, language and trajectories.
\newblock In {\em 2017 IEEE International Conference on Robotics and Automation
  (ICRA)},  2794--2801.
\newblock IEEE.

\bibitem[\protect\citeauthoryear{Szegedy \bgroup et al\mbox.\egroup
  }{2016}]{szegedy2016rethinking}
Szegedy, C.; Vanhoucke, V.; Ioffe, S.; Shlens, J.; and Wojna, Z.
\newblock 2016.
\newblock Rethinking the inception architecture for computer vision.
\newblock In {\em Proceedings of the IEEE conference on computer vision and
  pattern recognition},  2818--2826.

\bibitem[\protect\citeauthoryear{Wang \bgroup et al\mbox.\egroup
  }{2018}]{wang2018learning}
Wang, P.; Fu, Y.; Zhang, J.; Li, X.; and Lin, D.
\newblock 2018.
\newblock Learning urban community structures: A collective embedding
  perspective with periodic spatial-temporal mobility graphs.
\newblock {\em ACM Transactions on Intelligent Systems and Technology (TIST)}
  9(6):63.

\bibitem[\protect\citeauthoryear{Wang, Li, and
  Lazebnik}{2016}]{wang2016learning}
Wang, L.; Li, Y.; and Lazebnik, S.
\newblock 2016.
\newblock Learning deep structure-preserving image-text embeddings.
\newblock In {\em Proceedings of the IEEE conference on computer vision and
  pattern recognition},  5005--5013.

\bibitem[\protect\citeauthoryear{Yao \bgroup et al\mbox.\egroup
  }{2018}]{yao2018representing}
Yao, Z.; Fu, Y.; Liu, B.; Hu, W.; and Xiong, H.
\newblock 2018.
\newblock Representing urban functions through zone embedding with human
  mobility patterns.
\newblock In {\em IJCAI},  3919--3925.

\bibitem[\protect\citeauthoryear{Yuan, Zheng, and
  Xie}{2012}]{yuan2012discovering}
Yuan, J.; Zheng, Y.; and Xie, X.
\newblock 2012.
\newblock Discovering regions of different functions in a city using human
  mobility and pois.
\newblock In {\em Proceedings of the 18th ACM SIGKDD international conference
  on Knowledge discovery and data mining},  186--194.
\newblock ACM.

\bibitem[\protect\citeauthoryear{Zhang \bgroup et al\mbox.\egroup
  }{2017}]{zhang2017regions}
Zhang, C.; Zhang, K.; Yuan, Q.; Peng, H.; Zheng, Y.; Hanratty, T.; Wang, S.;
  and Han, J.
\newblock 2017.
\newblock Regions, periods, activities: Uncovering urban dynamics via
  cross-modal representation learning.
\newblock In {\em Proceedings of the 26th International Conference on World
  Wide Web},  361--370.
\newblock International World Wide Web Conferences Steering Committee.

\end{thebibliography}
\bibliographystyle{aaai}
\end{document}